# Failure Detection for Motion Prediction of Autonomous Driving: An Uncertainty Perspective

Wenbo Shao, Yanchao Xu*, Liang Peng, Jun Li, and Hong Wang*

*Abstract*— Motion prediction is essential for safe and efficient autonomous driving. However, the inexplicability and uncertainty of complex artificial intelligence models may lead to unpredictable failures of the motion prediction module, which may mislead the system to make unsafe decisions. Therefore, it is necessary to develop methods to guarantee reliable autonomous driving, where failure detection is a potential direction. Uncertainty estimates can be used to quantify the degree of confidence a model has in its predictions and may be valuable for failure detection. We propose a framework of failure detection for motion prediction from the uncertainty perspective, considering both motion uncertainty and model uncertainty, and formulate various uncertainty scores according to different prediction stages. The proposed approach is evaluated based on different motion prediction algorithms, uncertainty estimation methods, uncertainty scores, etc., and the results show that uncertainty is promising for failure detection for motion prediction but should be used with caution.

## I. INTRODUCTION

Motion prediction is a hot topic in mobile robot and autonomous vehicle communities, accurate prediction of the future motion of surrounding traffic participants is fundamental to robust and reliable decision-making. Artificial intelligence (AI), especially deep learning (DL), has been widely used in autonomous driving (AD) tasks by its advantages in dealing with complex problems. With the collection of large-scale data, the improvement of computing power and related algorithms, AI is expected to play a vital role in AD systems in the future [1].

However, although AI-based motion prediction has shown statistical performance advantages, it is difficult to avoid unpredictable failures due to the inherent inexplicability and insufficient reliability of DL models, which may cause serious AD accidents [2]. From the uncertainty perspective, motion prediction faces the dual challenge of uncertainty from the environment and the model. Drivers, pedestrians, etc. in the environment have uncertainty in their intentions and movements, which makes it difficult to accurately predict their future in all scenarios. Additionally, due to insufficient training data and training process, the model may experience serious performance degradation when faced with rare or unknown scenarios.

*Research supported by the National Science Foundation of China Project: U1964203 and 52072215, and the National Key R&D Program of China: 2020YFB1600303. (*Corresponding authors: Hong Wang, Yanchao Xu*)

Wenbo Shao, Liang Peng, Jun Li and Hong Wang are with School of Vehicle and Mobility, Tsinghua University, Beijing 100084, China. (e-mail: {swb19, peng-l20}@mails.tsinghua.edu.cn; lj19580324@126.com; hong_wang @mail.tsinghua.edu.cn)

Yanchao Xu is with School of Mechanical Engineering, Beijing Institute of Technology, Beijing 100081, China. (e-mail: 3120200410@bit.edu.cn)

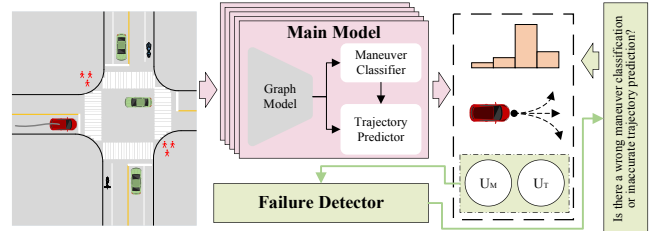

Fig. 1. Uncertainty-based failure detection for motion prediction. $U_M$, $U_T$ are the uncertainty scores (USs) extracted for maneuver classification and trajectory prediction, respectively.

The failure detection, isolation, and recovery mechanism is an effective way to solve the above problems [3]. Among them, the study of failure detection for AI models has attracted increasing interest, which is of critical significance for the development of reliable AD systems [4]. As shown in Fig. 1, using the information extracted from the main model, i.e. motion prediction model, a failure detector is built to identify maneuver classification errors and trajectory prediction errors. Uncertainty, as a measure of the confidence level of the model in its output, has been used by some researchers for failure detection in tasks such as semantic segmentation [5]. Our study exploits various uncertainties from motion prediction and explores their usefulness for failure detection.

In this work, we concentrate on failure detection for motion prediction from the uncertainty perspective. The main contributions are as follows:

- A framework of failure detection using uncertainty for motion prediction tasks, taking into account both motion uncertainty and model uncertainty.
- A series of USs for failure detection formulated for different motion prediction stages and algorithms.
- A detailed evaluation and comparison with multiple motion prediction algorithms, uncertainty estimation methods, and USs.

## II. RELATED WORK

### A. Motion Prediction and Motion Uncertainty Estimation

Traditional prediction methods predict the future motion of the target agent (TA) based on kinematic models [5], [6], but they only apply to short-term prediction. In recent years, DL-based methods [8]–[10] have demonstrated promising performance by simultaneously modeling TA's historical state, interaction, and other contexts in deep neural networks, a broader review can be found in [11]. Some studies model the network output as unimodal trajectories [12]–[14]. In addition, due to diverse intentions, uncertain behavior, and limited model inputs, TA's future motion presents multiple possibilities. Recently, increasing researchers and developers have paid attention to multimodal trajectory prediction, which is generally divided into two stages: maneuver classification and trajectory prediction. Some studies [15], [16] define maneuvers as specified behavior patterns, then train the maneuver classifier through supervised learning. For example, CS-LSTM [15] defines six maneuver modes for vehicles on highways, the predicted maneuvers provide guidance and interpretability for trajectory prediction. Additionally, some studies guide the model to learn implicit maneuvers through model design and training [17]–[19]. For example, Trajectron++ [18] adopts the conditional variational autoencoder (CVAE) and Gaussian mixture model (GMM) to realize multimodal probabilistic prediction.

### B. Model Uncertainty Estimation

The above multimodal prediction algorithms model the uncertainty in the TAs' movements. In addition, DL models have inherent uncertainty, generally called model uncertainty or epistemic uncertainty [20], it is difficult to ignore in the real world where there are distribution shifts or out-of-distribution data. Bayesian neural network (BNN) [21]–[23] is a representative method for estimating model uncertainty, in which Bayesian inference plays an important role. Methods such as Monte-Carlo dropout [24], [25] achieve approximate inference through sampling, and they further promote the generality and popularity of BNN. Besides, deep ensemble [26]–[28], as a simple and scalable method, has shown promising performance in model uncertainty estimation and thus has attracted many researchers and practitioners. As the representative method requiring only a single forward pass, evidential DL (EDL) [29] computes the uncertainty of the output distribution by modeling the prior distribution for the classification. Some studies [30], [31] try to improve the safety of AD by introducing model uncertainty estimation into the prediction or decision-making models. However, they are mainly verified based on data from the simulation, and there is a lack of comprehensive modeling and analysis of motion uncertainty and model uncertainty. This work simultaneously estimates two types of uncertainties in two stages of motion prediction and conducts detailed comparisons and analysis based on naturalistic driving datasets.

### C. Failure Detection for Autonomous Driving

Failure detection is attracting attention as a technology to achieve reliable AD. It uses the main model's input, internal features, or output to diagnose whether there is a failure. Learning-based approaches build a specialized model to act as the failure detector, and it identifies failures of the main model by using failure cases for supervised training [32]–[34] or estimating reconstruction errors [35]–[37]. In addition, uncertainty-based anomaly detection has attracted some interest, such as detecting misclassified or out-of-distribution examples through maximum softmax probabilities directly output by classification networks [38] or predictive entropy quantization taking into account model uncertainty [26]. However, to the best of our knowledge, most current research on failure detection for AD focuses on perception tasks, such as semantic segmentation, depth estimation, etc. [5], and failure detection for motion prediction models from the uncertainty perspective has been rarely discussed.

This work utilizes both motion uncertainty and model uncertainty, proposes USs for different stages of motion prediction, and investigates the performance of failure detection based on different scores.

### III. METHODOLOGY

#### A. Problem Setting

Motion prediction is a task that predicts TA's trajectory over a period of time in the future given input information. Assuming the current moment $t = 0$, the input information may include TA's historical state $\mathbf{S} = [s^{(-t_h+1)}, s^{(-t_h+2)}, \cdots, s^{(0)}]$ in the past $t_h$ timesteps, the historical state of TA's surrounding traffic participants, and other contexts $\mathbf{C}$ such as maps. Among them, $s^{(t)}$ may contain TA's information such as the position, speed, and category at $t$. The output is the predicted position $\hat{\mathbf{Y}}$ of TA in the future $t_f$ timesteps:

$$\hat{\mathbf{Y}} = f(\mathbf{S}, \mathbf{C}), \tag{1}$$

with $\hat{\mathbf{Y}} = [\hat{d}_1, \hat{d}_2, ..., \hat{d}_{t_f}]$ consisting of the $t_f$ predicted positions $\hat{d}_t$. For multimodal motion prediction, $\hat{\mathbf{Y}}$ contains predicted trajectories under multiple maneuvers.

Failure detection for motion prediction aims to identify potential motion prediction failures by monitoring model's state, where failures may exist in the form of maneuver misclassification or excessive error of predicted trajectories. Uncertainty, as the measure of TA's behavior or model state, reflects the model's confidence in its particular output and thus has the potential to diagnose potential prediction failures. This work proposes to detect the performance degradation of motion prediction models, i.e. the decrease in the prediction accuracy, by quantifying the USs.

#### B. Motion Prediction with Motion Uncertainty Estimation

Due to the unavailability of TA's actual intentions and the randomness of its behavior, it may have multiple possible future trajectories. GRIP++ is an enhanced graph-based interaction-aware trajectory prediction algorithm, it models inter-agent interactions and temporal features but only predicts future trajectories in a single mode. As shown in Fig. 2, we propose a prediction model that includes a maneuver classification module to improve the authenticity and usability of the prediction results, which is called GRIP+++.

We focus on two-stage tasks in the proposed method: maneuver classification and maneuver-based trajectory prediction. In the maneuver classification stage, given the TA's historical state and scene context, the feature $G$ is extracted through the graph convolutional model (GCN), which includes the processing of fixed and trainable graphs. Then TA's maneuver $z \sim \mathrm{P}(z|G)$ is inferred by multi-layer perceptron (MLP) models, where $z \in \{1, 2, ..., Z\}$ represents one of the defined maneuver modes. In CS-LSTM [15], the modes are divided into three types of lateral maneuvers and two types of longitudinal maneuvers, but they are only applicable to vehicles driving on highways, we define a common set of maneuver modes suitable for various scenarios. Specifically, TA's maneuvers are divided into four categories according to their movement direction and speed: going straight, turning left, turning right, and stopping. In the network, we adopt the softmax function for probabilistic maneuver classification.

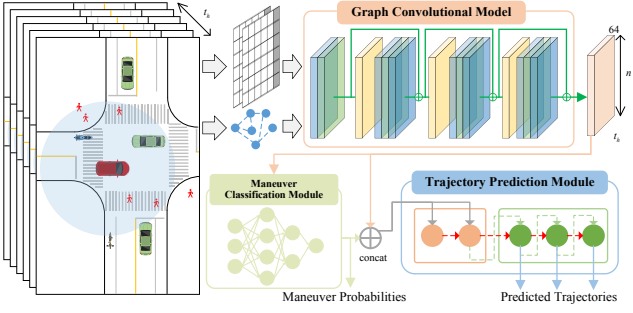

Fig. 2. The architecture of GRIP+++.

The maneuver-based trajectory prediction module consists of seq2seq networks taking the concatenation of the graph feature $G$ and the feature vector transformed by the maneuver $z$ as input, and outputs the future trajectory $\hat{\mathbf{Y}}_z$ under the maneuver $z$.

To compare the generality of uncertainty-based failure detection in different motion prediction mechanisms, we employ another two classes of typical prediction algorithms. Firstly, we focus on multimodal trajectory prediction based on the generative model, so we adopt Trajectron++ [18], it utilizes the CVAE-based latent network framework to model multimodal future trajectories, where the discrete categorical latent variable $z$ encodes high-level behavior patterns:

$$\hat{\mathbf{Y}} \sim \mathrm{P}(\hat{\mathbf{Y}}|\mathbf{S},\mathbf{C}) = \sum_{z \in \{1,2,\ldots,Z\}} \mathrm{P}_{\psi}(\hat{\mathbf{Y}}|\mathbf{S},\mathbf{C},z)\mathrm{P}_{\theta}(z|\mathbf{S},\mathbf{C}), \quad (2)$$

where $\theta$, $\psi$ are deep neural network parameters.

Furthermore, we use PGP [16] as a comparison, it is a multimodal trajectory prediction method combining graph traversal, latent vector sampling, and clustering. It models discrete policy for graph traversal by representing HD maps as lane graphs, and implements diverse trajectories prediction combined with a random sampling of latent vectors for longitudinal variability. Furthermore, it uses K-means clustering to obtain $Z$ predictive trajectories. With its clever design, PGP achieved state-of-the-art results on almost all metrics of the nuScenes leaderboard when proposed.

### C. Model Uncertainty Estimation

As mentioned above, deep ensemble has certain advantages in model uncertainty estimation, so we design a prediction approach that simultaneously integrates model uncertainty and motion uncertainty estimation based on it. Specifically, we use random initialization of the model parameters and random shuffling of the training data to train $K$ homogeneous and heterogeneous models, then estimate uncertainty based on the $K$ set of output $\hat{\mathbf{Y}}^k$, $k \in \{1,2,\ldots,K\}$.

In addition, EDL, as a method to capture multiclass uncertainties with low computational cost, is also exploited to estimate the model uncertainty of the maneuver classification module. Specifically, the Dirichlet distribution is considered the prior distribution for the classification:

$$D(\mathrm{P}|\boldsymbol{\alpha}) = \begin{cases} \dfrac{1}{B(\boldsymbol{\alpha})} \prod_{z=1}^{Z} \mathrm{P}_z^{\alpha_z - 1} & \text{for } \mathrm{P} \in \mathcal{S}_Z \\ 0 & \text{otherwise} \end{cases}, \quad (3)$$

where $\boldsymbol{\alpha} = [\alpha_1, \ldots, \alpha_Z]$ are the distribution parameters, $e_z = \alpha_z - 1$ is the evidence, and $\mathcal{S}_Z$ is the $Z$-dimensional unit simplex.

### D. Uncertainty Scores Design

In our work, different USs are proposed for failure detection. Considering the different problem forms of maneuver classification and trajectory prediction tasks, we formulate corresponding USs for both.

For maneuver classification task combined with deep ensemble, we formulate the following USs referring to the definition in [39]:

Total entropy (TE) for maneuver classification is quantified to represent the total uncertainty considering both model uncertainty and the motion uncertainty:

$$\mathrm{TE} = \mathcal{H}\left[\mathbb{E}_{\mathrm{P}(\boldsymbol{\theta}|\mathcal{D})}\left[\mathrm{P}(z|\mathbf{S},\mathbf{C},\boldsymbol{\theta})\right]\right] = \mathcal{H}\left[\frac{1}{K}\sum_{k=1}^{K}\mathrm{P}(z|\mathbf{S},\mathbf{C},\boldsymbol{\theta}_k)\right], \quad (4)$$

where $\boldsymbol{\theta}_k$ are the parameters of the $kth$ model of deep ensemble, $\mathcal{H}$ represents the formula for calculating entropy, $\mathcal{D}$ represents the training set.

Data entropy (DE) for maneuver classification is quantified to represent the average of data uncertainty from different models. The larger the value, the higher the motion uncertainty estimated by deep ensemble-prediction models.

$$\mathrm{DE} = \mathbb{E}_{\mathrm{P}(\boldsymbol{\theta}|\mathcal{D})}\left[\mathcal{H}(z|\mathbf{S},\mathbf{C},\boldsymbol{\theta})\right] = \frac{1}{K}\sum_{k=1}^{K}\mathcal{H}\left[\mathrm{P}(z|\mathbf{S},\mathbf{C},\boldsymbol{\theta}_k)\right], \quad (5)$$

Mutual Information (MI) is quantified to represent the model uncertainty. As it increases, the degree of difference between the prediction results of multiple models increases, which to a certain extent reflects the reduction of the confidence of the models in their classification results.

$$\mathrm{MI} = \mathcal{I}\left[z, \boldsymbol{\theta} | \mathbf{S}, \mathbf{C}, \mathcal{D}\right] = \mathrm{TE} - \mathrm{DE}, \quad (6)$$

The maximum predicted probability [40] is also considered and its inverse (negative maximum softmax probability, NMaP) is calculated as an US.

As for the EDL-based method, the above-discussed types of USs are also quantified for comparison, and their formulas are derived according to (3) - (6). Additionally, we consider the metrics suggested in [29]:

$$\mathrm{u} = \frac{Z}{\sum_{z=1}^{Z} \alpha_z}, \quad (7)$$

Trajectory prediction involves multiple trajectories output by one or more models, where each trajectory contains position information for multiple future moments. Referring to the usual error metrics, average displacement error (ADE) and final displacement error (FDE) [8], [12], [18], we define two basic USs, average predictive entropy (APE) and final predictive entropy (FPE):

$$\mathrm{APE} = \frac{1}{t_f}\sum_{i=1}^{t_f} \mathcal{H}\left[\hat{d}_{t_i}\right] = \frac{1}{t_f}\sum_{i=1}^{t_f}\left[(\ln 2\pi + 1) + \frac{1}{2}\ln\left|\hat{\Sigma}_{t_i}\right|\right], \quad (8)$$

$$\mathrm{FPE} = \mathcal{H}\left[\hat{d}_{t_f}\right] = (\ln 2\pi + 1) + \frac{1}{2}\ln\left|\hat{\Sigma}_{t_f}\right|, \quad (9)$$

where for different predicted trajectories for one input, the predicted position $\hat{d}_t$ at the same time is assumed to follow a bivariate Gaussian distribution. Based on the above two basic metrics, different types of USs are defined according to the source of different predicted trajectories (such as different sub-models, different maneuvers, or both), which represent model uncertainty, motion uncertainty, or total uncertainty.

## IV. EXPERIMENTS

### A. Experimental Setup

*1) Model Implementation:* For the training of GRIP+++, inspired by [15], we adopt a two-stage training approach. In the first stage, we focus on improving the trajectory prediction accuracy under the real maneuver, by training the model as a regression task at each time:

$$L_{reg} = \frac{1}{t_f}\sum_{t=1}^{t_f}\left\|\hat{\mathbf{Y}}_{t,z} - \mathbf{Y}_t\right\|, \quad (10)$$

where $\hat{\mathbf{Y}}_{t,z}$ and $\mathbf{Y}_t$ are predicted positions for true maneuver $z$ and ground truth at time $t$ respectively.

In the second stage, we additionally consider the loss of maneuver classification by adding the cross-entropy loss:

$$L = L_{reg} + \lambda L_{man}, \quad (11)$$

where $L_{man} = -\log(P(z|\mathbf{S},\mathbf{C}))$, $\lambda$ is the weighting factor, and $z$ is the true maneuver label. Besides, in the implementation of GRIP+++, the trajectories are sampled at 2Hz, with an observation length of 3s and a prediction horizon of 3s.

As for the implementation of Trajectron++ [18] and PGP [16], we follow their original model design and training scheme. For deep ensemble, we set $K = 5$, a scheme considered cost-controllable and sufficiently efficient. To achieve EDL, referring to [29], we incorporate a Kullback-Leibler (KL) divergence term into our loss function to avoid unnecessary uncertainty reduction.

*2) Dataset:* The models are trained and tested on real traffic datasets. Specifically, GRIP+++ and its failure detectors are trained on the SinD training set and tested on the SinD test set and INTERACTION, respectively. The SinD [41] dataset consists of 13248 recorded trajectories from a signalized intersection. The traffic participants include cars, trucks, buses, tricycles, bikes, motorcycles, and pedestrians. The INTERACTION [42] dataset contains data collected in four categories of scenarios, where we adopt the TC_intersection_VA (VA) subset collected from a signaled intersection in Bulgaria and the USA_intersection_GL (GL) subset collected from a signaled intersection in the USA. For each type of agent, based on variation in its position, speed, and heading angle at every 0.5-second interval, we apply a specific threshold to annotate its maneuver mode offline, including going straight, turning left, turning right, and stopping. Trajectron++, PGP, and their failure detection experiments are carried out on the nuScenes. NuScenes [43] is a large-scale self-driving car dataset with 1000 scenes, each scene contains 20s object annotations and HD semantic maps.

*3) Evaluation methodology:* We set the evaluation methodology separately for the failure detection for the two-stage prediction task. For the maneuver classification task, a good failure detector is considered to assign higher USs to misclassified cases. Therefore, we adopt the area under the receiver operating characteristic curve (AUROC) as the basic evaluation metric, where the misclassified data are taken as positive samples. However, AUROC does not reflect the impact of the addition of the failure detection module on the performance of the original prediction algorithm. Therefore, we also plot the cut-off curve to evaluate the average accuracy of the remaining data after filtering out a certain percentage of data in descending order of uncertainty. The area under the cut-off curve (AUCOC) is regarded as an overall evaluation of the prediction model with the failure detector, with a larger value indicating better performance. For the trajectory prediction task, AUROC is not suitable, we use the cut-off curve as the evaluation methodology. Unlike maneuver classification, the curve here is drawn by calculating the average prediction error of the remaining data, so a smaller AUCOC represents better performance.

### B. Failure Detection for Maneuver Classification

Regarding failure detection for maneuver classification, we set up several experiments to answer the following questions.

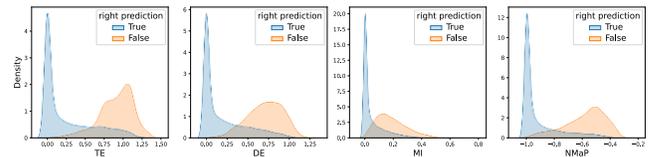

Fig. 3. Uncertainty distribution for correctly classified and misclassified samples. Experimental results of deep ensemble-based GRIP+++ on SinD.

**How different are the distributions of USs for correct and misclassified cases?** An effective uncertainty-based failure detector is built on the assumption that the US level has a strong correlation with the prediction correctness. As in Fig. 3, the USs of the correctly predicted maneuvers are generally relatively low, while the incorrectly predicted cases generally have high USs. Meanwhile, there is a relatively obvious separation between the two distributions, especially for TA, DA, and NMaP. Therefore, it is preliminarily inferred that the USs have the potential for failure detection.

**Comparison of different USs for failure detection?** As indicated previously, in the deep ensemble-based maneuver classification network, we can extract various USs, here we set up experiments to compare the effects of different scores as the reference for failure detection. The second row of TABLE I shows the results, NMaP, TE, and DE achieve better failure detection performance when used as USs, where the total uncertainty considering both motion and model uncertainty is slightly better than the motion uncertainty alone. NMaP is relatively simple to calculate and has a strong detection ability. Furthermore, although MI, which represents the model uncertainty, reflects the reduced confidence of the model when faced with new scenarios (as in TABLE II), its performance is not good enough when used alone for failure detection. In Fig. 4, the cut-off curve and AUCOC corresponding to different USs are further compared. They have a great advantage over the random filtering method and are close to the optimal situation, and the relative relationship between different USs is consistent with TABLE I.

TABLE I. AUROC(↑) FOR MANEUVER CLASSIFICATION STAGE OF GRIP+++ ON SIND. MODEL i IS THE RESULT OF THE ITH MODEL FROM DEEP ENSEMBLE

|          | TE    | DE    | MI    | NMaP  | u     |
|----------|-------|-------|-------|-------|-------|
| Ensemble | 0.911 | 0.903 | 0.864 | **0.918** | -     |
| Model 1  | -     | 0.871 | -     | 0.867 | -     |
| Model 2  | -     | 0.868 | -     | 0.864 | -     |
| Model 3  | -     | 0.871 | -     | 0.867 | -     |
| Model 4  | -     | 0.868 | -     | 0.864 | -     |
| Model 5  | -     | 0.863 | -     | 0.858 | -     |
| EDL      | **0.912** | **0.909** | **0.911** | 0.912 | 0.910 |

TABLE II. AVERAGE UNCERTAINTY OBTAINED BY DEEP ENSEMBLE-BASED GRIP+++ TRAINED ON SINK, AND TESTED ON IN-DISTRIBUTION DATA (SINK) AND OUT-OF-DISTRIBUTION DATA (VA), RESPECTIVELY

|      | TE    | DE    | MI    | NMaP   |
|------|-------|-------|-------|--------|
| SinD | 0.318 | 0.250 | 0.068 | -0.877 |
| VA   | 0.299 | 0.199 | 0.100 | -0.880 |
| GL   | 0.429 | 0.280 | 0.148 | -0.826 |

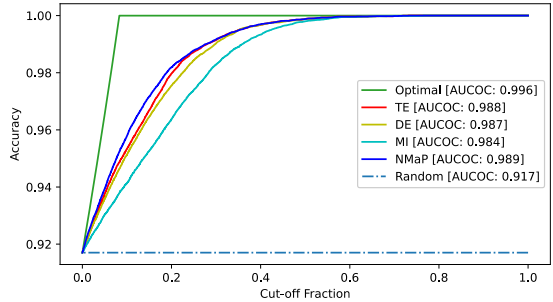

Fig. 4. Cut-off curves and AUCOC (↑) for maneuver classification stage of deep ensemble-based GRIP+++ on SinD. The optimal curve is drawn by directly using the classification error as a filtering reference; the random curve is drawn by filtering the data in random order.

TABLE III. AUCOC (↑) FOR MANEUVER CLASSIFICATION STAGE OF GRIP+++ ON SINK

|          | TE    | DE    | MI    | NMaP  | u     |
|----------|-------|-------|-------|-------|-------|
| Ensemble | **0.988** | **0.987** | **0.984** | **0.989** | -     |
| Model 1  | -     | 0.981 | -     | 0.982 | -     |
| Model 2  | -     | 0.980 | -     | 0.981 | -     |
| Model 3  | -     | 0.981 | -     | 0.982 | -     |
| Model 4  | -     | 0.980 | -     | 0.980 | -     |
| Model 5  | -     | 0.979 | -     | 0.979 | -     |
| EDL      | 0.978 | 0.978 | 0.978 | 0.978 | 0.978 |

**USs based on deep ensemble vs. USs based on a single model?** Here, we obtain DE and NMaP from the single model in deep ensemble, and they are further used for failure detection for the maneuver classification module of the corresponding model. From the comparison of rows 2-7 of TABLE I, although the USs extracted from the single model have a certain failure detection ability, they are not as good as the failure detector based on deep ensemble. Moreover, the DE extracted from the motion prediction model integrated with the model uncertainty estimation has better failure detection performance than the DE extracted from the single model. In addition, it is also concluded from the comparison of rows 2-7 in TABLE III that the introduction of deep ensemble is beneficial to improve the maneuver classification performance combined with failure detector filtering.

**How well do the EDL-based USs perform?** As a comparison, we employ EDL to extract USs and evaluate their performance for failure detection. TABLE I shows that using the USs extracted by EDL as references for the failure detector achieves comparable results to deep ensemble. However, TABLE III presents that the overall accuracy after filtering the data by the EDL-based USs is not high. One possible reason is that the regularization term added by EDL during the training process causes a drop in the prediction performance of the main model, which in turn weakens the effect of motion prediction with failure detection.

*C. Failure Detection for Trajectory Prediction*

As for failure detection for trajectory prediction, we design some experiments to answer the following questions.

**How well does the failure detector based on USs from multiple trajectories perform?** For the prediction error, considering the $K$ predicted trajectories under the real maneuver $z$, we calculate the minimum (minADE$_z$, minFDE$_z$) and mean (meanADE$_z$, meanFDE$_z$) of the errors of the $K$ trajectories, and the error of their average trajectory (ADE$_{z, avg}$, FDE$_{z, avg}$). We calculate APE$_z$ and FPE$_z$ of the above $K$ trajectories to estimate the predictive uncertainty. As a comparison, we calculate the uncertainty of the average trajectories of $K$ models in different maneuvers (APE$_{avg}$, FPE$_{avg}$), which to some extent represent the motion uncertainty. In TABLE IV, each column represents an error metric and each row represents the corresponding US used for failure detection (except rows 1-3). By comparing rows 2-5 and 11-14, APE$_z$ and FPE$_z$ achieve better results than APE$_{avg}$ and FPE$_{avg}$, showing that model uncertainty has a stronger failure detection potential than motion uncertainty. There are still some failures that are difficult to detect. This limitation is caused by the gap between the complex real world and the assumptions of the uncertainty estimation method, and the lack of effective guidance during obtaining uncertainty is also one of the reasons. Our future work will try to further improve the performance of failure detection from the above aspects.

**Are the USs extracted in the maneuver classification stage applicable to the trajectory prediction stage?** Theoretically, the USs obtained in the maneuver classification stage represent the confidence of the model in the current scene, so it may be suitable for failure detection in the trajectory prediction stage. We conduct some experiments to explore this question, the results are recorded in rows 6-9 of the two sub-tables of TABLE IV. Compared with the above trajectory USs, the uncertainty extracted in the maneuver classification stage has limited potential for detecting high-error trajectories. One of the possible reasons is that the USs calculated directly based on the trajectories imply the consideration of information such as the velocity and acceleration of the object, thus having a greater correlation with the trajectory error.

**How is the failure detection generalizing to scenarios with larger distributional shifts?** Here, we use the VA and GL datasets to test the model trained based on SinD. Due to the differences in the location, period, and type of intersection of the data source, the test sets have large distributional shifts with the training set, which causes the generalization challenge of the DL-based prediction model. The results are shown in TABLE V and TABLE VI. Compared with TABLE I, III, and IV, the prediction accuracy of the main model deteriorates due to larger distribution offset, that is, it will face larger failure probability. The proposed failure detector can detect failure cases by estimating uncertainty, which is convenient for AD to deal with these high-risk scenarios.

TABLE IV. AUCOC (↓)/IMPROVEMENT SCORE (IS)[1](↑) FOR TRAJECTORY PREDICTION STAGE OF GRIP+++ WITH DEEP ENSEMBLE ON SIND

|  | $minADE_z$ | $meanADE_z$ | $ADE_{z, avg}$ |
|---|---|---|---|
| **Optimal** | 0.066 | 0.096 | 0.088 |
| **Random** | 0.259 | 0.345 | 0.330 |
| **$APE_z$** | **0.119/0.725** | **0.143/0.813** | **0.139/0.790** |
| **$APE_{avg}$** | 0.136/0.636 | 0.172/0.694 | 0.166/0.677 |
| **TE** | 0.170/0.459 | 0.228/0.469 | 0.218/0.464 |
| **DE** | 0.170/0.457 | 0.229/0.466 | 0.218/0.462 |
| **MI** | 0.169/0.462 | 0.227/0.476 | 0.216/0.470 |
| **NMaP** | 0.170/0.461 | 0.228/0.472 | 0.217/0.467 |
|  | $minFDE_z$ | $meanFDE_z$ | $FDE_{z, avg}$ |
| **Optimal** | 0.114 | 0.182 | 0.164 |
| **Random** | 0.522 | 0.718 | 0.686 |
| **$FPE_z$** | **0.249/0.670** | **0.301/0.779** | **0.293/0.754** |
| **$FPE_{avg}$** | 0.278/0.599 | 0.358/0.672 | 0.345/0.654 |
| **TE** | 0.361/0.395 | 0.493/0.420 | 0.471/0.413 |
| **DE** | 0.362/0.393 | 0.494/0.417 | 0.472/0.410 |
| **MI** | 0.359/0.400 | 0.489/0.428 | 0.467/0.420 |
| **NMaP** | 0.360/0.397 | 0.491/0.423 | 0.497/0.416 |

TABLE V. AUROC(↑)/AUCOC(↑) FOR MANEUVER CLASSIFICATION STAGE OF GRIP+++ WITH DEEP ENSEMBLE, MODEL IS TRAINED ON SIND AND TESTED ON VA/GL

|  | TE | DE | MI | NMaP |
|---|---|---|---|---|
| **VA** | 0.914/0.978 | 0.915/0.978 | 0.863/0.971 | 0.912/0.978 |
| **GL** | 0.792/0.909 | 0.793/0.910 | 0.752/0.900 | 0.793/0.909 |

TABLE VI. $AUCOC_{UNCERTAINTY}$ (↓)/IS(↑) FOR TRAJECTORY PREDICTION STAGE OF GRIP+++ WITH DEEP ENSEMBLE, MODEL IS TRAINED ON SIND AND TESTED ON VA/GL

|  | $minADE_z$ | $meanADE_z$ | $ADE_{z, avg}$ |
|---|---|---|---|
| **$APE_z$ (VA)** | 0.210/0.656 | 0.238/0.744 | 0.234/0.730 |
| **$APE_z$ (GL)** | 0.413/0.617 | 0.505/0.746 | 0.492/0.709 |
|  | $minFDE_z$ | $meanFDE_z$ | $FDE_{z, avg}$ |
| **$FPE_z$ (VA)** | 0.491/0.601 | 0.550/0.699 | 0.543/0.686 |
| **$FPE_z$ (GL)** | 1.002/0.580 | 1.209/0.721 | 1.183/0.687 |

TABLE VII. $AUCOC_{OPTIMAL}/AUCOC_{UNCERTAINTY}$ (↓)/$AUCOC_{RANDOM}$/IS(↑) FOR TRAJECTRON++ ON NUSCENES

|  | Single model | Deep ensemble |
|---|---|---|
| **(mean)minADE** | 0.088/0.167/0.378/0.730 | 0.096/0.160/0.384/0.778 |
| **(mean)minFDE** | 0.132/0.308/0.689/0.683 | 0.151/0.293/0.702/0.742 |
| **(mean)meanADE** | 0.322/0.386/1.045/0.912 | 0.339/0.394/1.040/0.922 |
| **(mean)meanFDE** | 0.608/0.754/2.096/0.902 | 0.637/0.763/2.082/0.913 |
| **minminADE** | - | 0.055/0.112/0.234/0.682 |
| **meanmaxpADE** | - | 0.181/0.280/0.801/0.841 |

TABLE VIII. $AUCOC_{OPTIMAL}/AUCOC_{UNCERTAINTY}$(↓)/$AUCOC_{RANDOM}$/IS(↑) FOR PGP ON NUSCENES, UC MEANS UNIFIED CLUSTERING

|  | Single model | Deep ensemble |
|---|---|---|
| **(mean)minADE** | 0.498/0.837/0.945/0.242 | 0.529/0.832/0.945/0.271 |
| **(mean)minFDE** | 0.623/1.273/1.554/0.302 | 0.747/1.249/1.548/0.373 |
| **minminADE** | - | 0.367/0.628/0.708/0.234 |
| **meanmaxpADE** | - | 1.538/2.497/3.115/0.392 |
| **minADE (uc)** | - | 0.488/0.797/0.908/0.264 |
| **minFDE (uc)** | - | 0.612/0.181/1.466/0.333 |

**How well does uncertainty-based failure detection perform in generative model-based trajectory prediction?** We adopt Trajectron++ combined with deep ensemble to extract multiple USs as failure detection references. The results of this investigation are provided in TABLE VII, where minADE/minFDE/meanADE/meanFDE for a single model is calculated based on the 10 trajectories predicted by the model, and the corresponding USs for failure detection are APE/FPE/APE/FPE obtained from the 10 trajectories. In contrast, meanminADE/meanminFDE/meanmeanADE/meanmeanFDE/minminADE/meanmaxpADE for deep ensemble are calculated based on 50 trajectories from all 5 ensemble models, where the first operator (mean/min) is for different sub-models and the second operator (mean/min/maxp) is for different maneuvers from each model's output. The corresponding USs for failure detection are meanAPE/meanFPE/meanAPE/meanFPE/$APE_{all}$/$APE_{maxp}$, where meanAPE/meanFPE are obtained by averaging APE/FPE from 5 sub-models, $APE_{all}$ is calculated based on all 50 trajectories, $APE_{maxp}$ is calculated according to the trajectory of maximum probability of each model. The results show promising performance of the uncertainty-based failure detector.

**Can the above uncertainty-based failure detection be directly applied to any trajectory prediction algorithms?** In addition to the typical deep neural network architecture and modules, existing trajectory prediction algorithms may use various tricks, which may directly affect the USs extracted from the output trajectories. We conduct some exploratory experiments with PGP, a high-performance prediction algorithm integrating special tricks including traversal, sampling, and clustering, to analyze the performance of applying the USs obtained from the output trajectories for failure detection. In addition, we apply deep ensemble to consider model uncertainty. From the evaluation results in TABLE VIII, we conclude that the performance of direct uncertainty quantification based on output results is not very outstanding. Possible reasons include operations such as sampling latent vectors from an unconstrained normal distribution and clustering. This result reminds us that it is necessary to improve uncertainty estimation methods and scores according to the prediction algorithms' characteristics. For example, we propose a framework for unified clustering based on the outputs of all sub-models of the deep ensemble, the results in the last two rows of TABLE VIII show some improvement over the original single model in both trajectory prediction and failure detection performance.

## V. CONCLUSION

In this work, we propose a framework to detect motion prediction failures from the uncertainty perspective. We divide motion prediction tasks into two stages, maneuver classification and maneuver-based trajectory prediction, and formulate corresponding USs for failure detection, where motion uncertainty and model uncertainty are both discussed. Our experiments cover the comparison of different prediction tasks, multiple prediction algorithms, different uncertainty estimation methods, and various USs. Finally, we observe that uncertainty quantification is promising for failure detection for motion prediction, with the potential to generalize to environments with larger distributional shifts. However, it is also necessary to conduct targeted discussions and designs for different prediction algorithms. Our future work will focus on the integration of the proposed method with safety decision-making for AD, and its implementation and validation on physical vehicle platforms.

---

[1] IS is calculated by $(AUCOC_{random} - AUCOC_{uncertainty})/(AUCOC_{random} - AUCOC_{optimal})$, where $AUCOC_{random}$, $AUCOC_{optimal}$, and $AUCOC_{uncertainty}$ represent the AUCOC based on the optimal sorting, the random sorting, and the USs-based sorting, respectively.